\documentclass{article}

\usepackage[final]{neurips_2021}
\usepackage[utf8]{inputenc}

\usepackage{textcomp}
\usepackage{titlesec}
\usepackage{array}
\usepackage{afterpage}

\usepackage{caption}

\usepackage{hyperref}
\usepackage{url}
\usepackage{subcaption}
\usepackage{natbib}
\setcitestyle{numbers,square}
\usepackage{graphicx}
\usepackage{fancyvrb}
\usepackage{booktabs}

\usepackage{enumitem}

\usepackage{xcolor}
\definecolor{deepblue}{RGB}{0,0,139}

\hypersetup{
    colorlinks=true,
    citecolor=deepblue,
    linkcolor=deepblue,
    filecolor=deepblue,      
    urlcolor=deepblue,
}

\begin{document}

\title{Revolutionizing Retrieval-Augmented Generation\\ with Enhanced PDF Structure Recognition}

\author{Demiao LIN\\\href{http://www.chatdoc.com}{chatdoc.com}}

\maketitle

\begin{abstract}

With the rapid development of Large Language Models (LLMs), Retrieval-Augmented Generation (RAG) has become a predominant method in the field of professional knowledge-based question answering. Presently, major foundation model companies have opened up Embedding and Chat API interfaces, and frameworks like LangChain have already integrated the RAG process. It appears that the key models and steps in RAG have been resolved, leading to the question: are professional knowledge QA systems now approaching perfection? 
This article discovers that current primary methods depend on the premise of accessing high-quality text corpora. However, since professional documents are mainly stored in PDFs, the low accuracy of PDF parsing significantly impacts the effectiveness of professional knowledge-based QA. We conducted an empirical RAG experiment across hundreds of questions from the corresponding real-world professional documents. The results show that, ChatDOC (\href{http://www.chatdoc.com}{chatdoc.com}), a RAG system equipped with a panoptic and pinpoint PDF parser, retrieves more accurate and complete segments, and thus better answers. Empirical experiments show that ChatDOC is superior to baseline on nearly 47\% of questions, ties for 38\% of cases, and falls short on only 15\% of cases. It shows that we may revolutionize RAG with enhanced PDF structure recognition.
\end{abstract}

\section{Introduction}

Large language models (LLM) are trained on data that predominantly come from publicly available internet sources, including web pages, books, news, and dialogue texts. It means that LLMs primarily rely on internet sources as their training data, which are vast, diverse, and easily accessible, supporting them to scale up their capabilities. However, in vertical applications, professional tasks require LLMs to utilize domain knowledge, which  unfortunately is private, and not part of their pre-training data.

\begin{figure}[t]
\centering
\includegraphics[width=.85\textwidth]{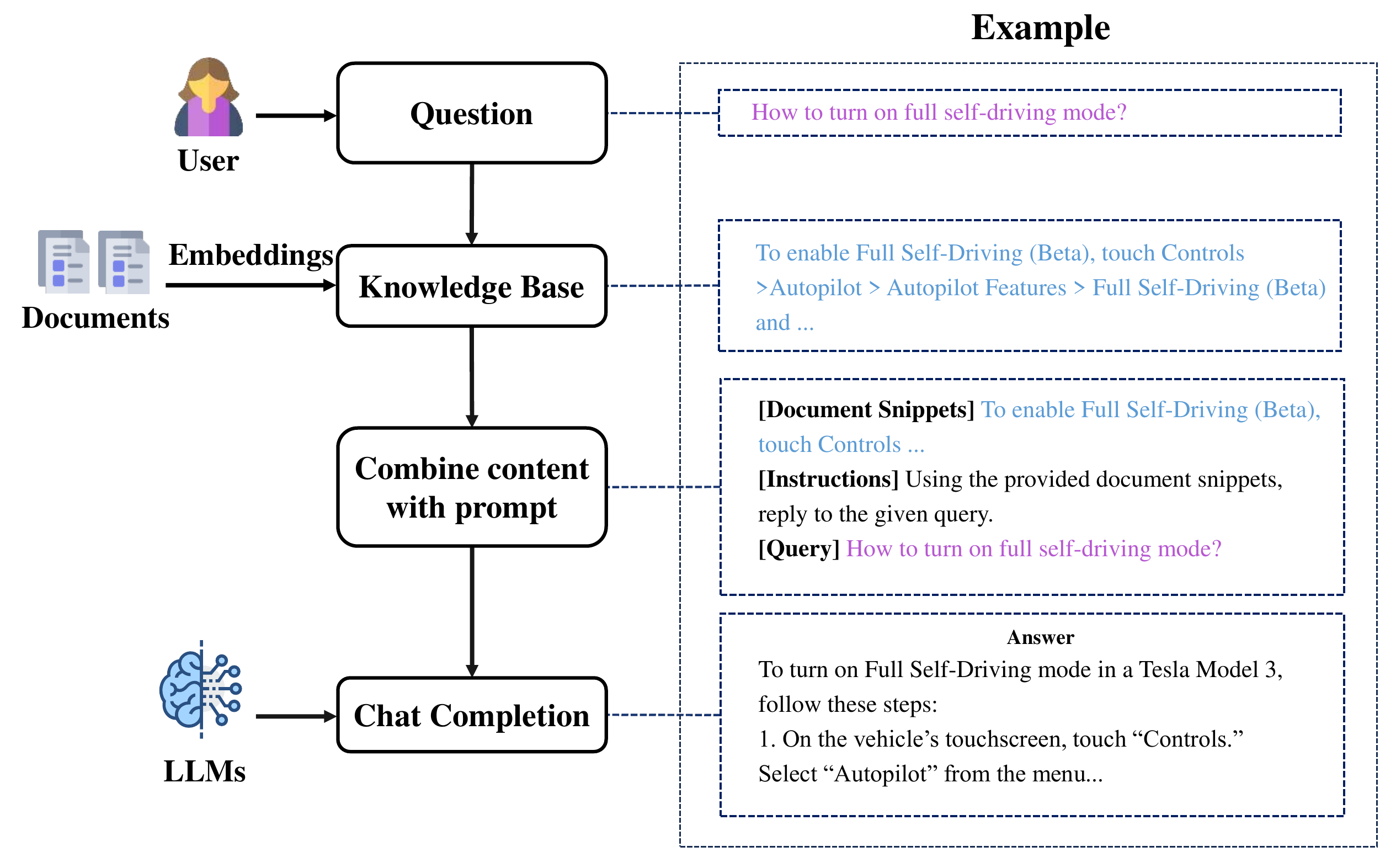}
\caption{The workflow of Retrieval-Augmented Generation (RAG).}
\label{fig:ragworkflow}
\end{figure}

A popular approach to equip LLM with domain knowledge is Retrieval-Augmented Generation (RAG). RAG framework answers a question in four steps: the user proposes a query, the system retrieves the relevant content from private knowledge bases, combines it with the user query as context, and finally asks the LLM to generate an answer.
This is illustrated in \ref{fig:ragworkflow} with a simple example. This process mirrors the typical cognitive process of encountering a problem, including consulting relevant references and subsequently deriving an answer. In this framework, the pivotal component is the accurate retrieval of pertinent information, which is critical for the efficacy of the RAG model.

\begin{figure}[t]
\centering
\includegraphics[width=.85\textwidth]{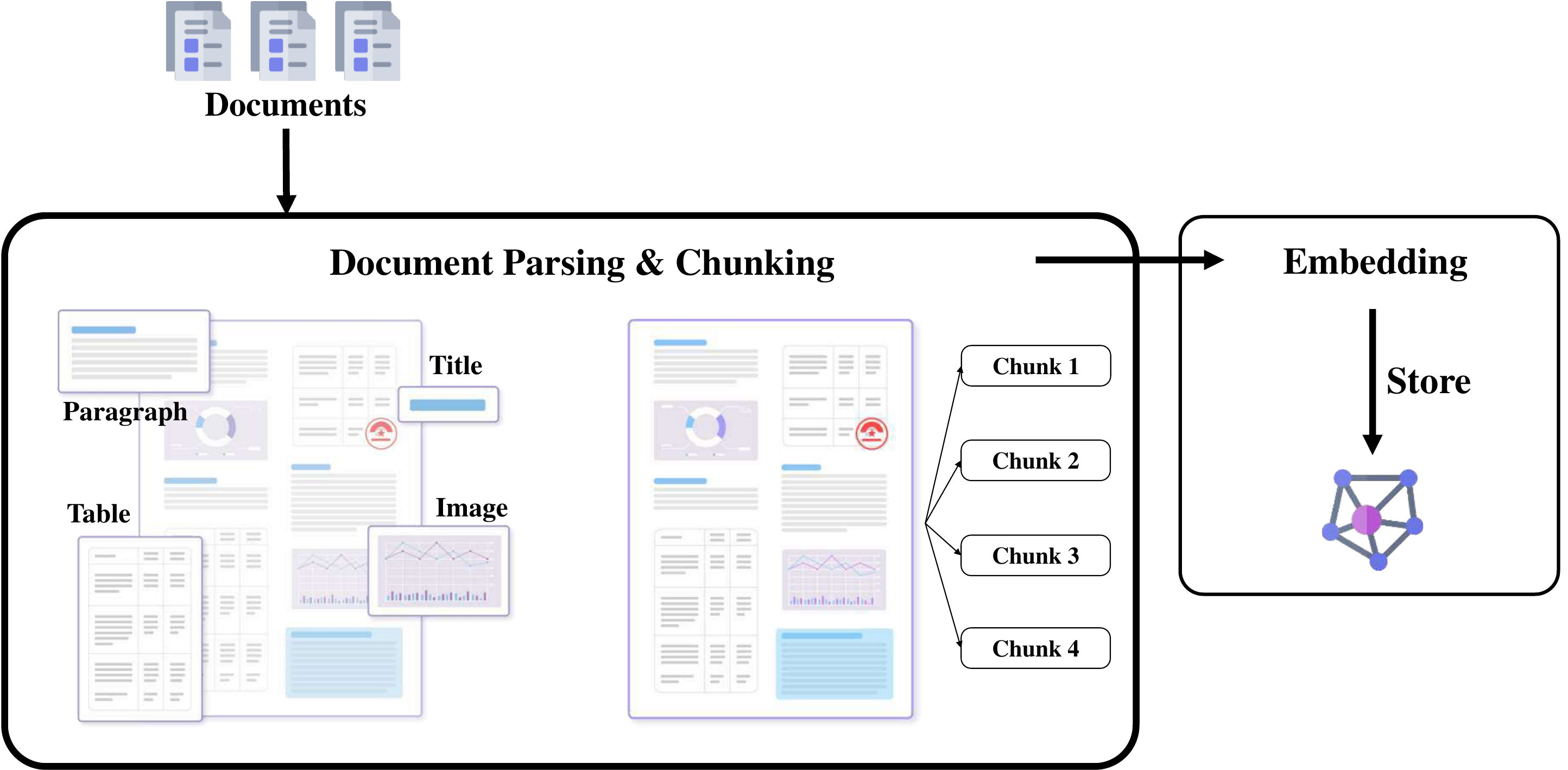}
\caption{The process of converting PDFs into retrievable contents. }
\label{fig:pdfconvert}
\end{figure}

However, the process of retrieval from PDF files is fraught with challenges. Common issues include inaccuracies in text extraction and disarray in the row-column relationships of tables inside PDF files. Thus, before RAG, we need to convert large documents into retrievable content. The conversion involves several steps, as shown in \ref{fig:pdfconvert}:
\begin{itemize}[leftmargin=0.5cm]
  \item Document Parsing \& Chunking. It involves extracting paragraphs, tables, and other content blocks, then dividing the extracted content into chunks for subsequent retrieval.
  \item Embedding. It transforms text chunks into real-valued vectors and stores them in a database.
\end{itemize}
Since each of these steps can lead to information loss, the compounded losses can significantly impact the effectiveness of RAG’s responses. 

This paper primarily addresses the question of whether the quality of PDF parsing and chunking affects the outcomes of RAG. We will explore the challenges, methodologies, and real-world case studies pertaining to this issue. It will include an examination of two types of methods in this field, namely rule-based and deep learning-based methods, followed by empirical evaluations of their efficacy through practical examples.

\section{PDF Parsing \& Chunking}

\subsection{Challenges and Methods Overview}

\begin{figure}[t]
\centering
\includegraphics[width=.75\textwidth]{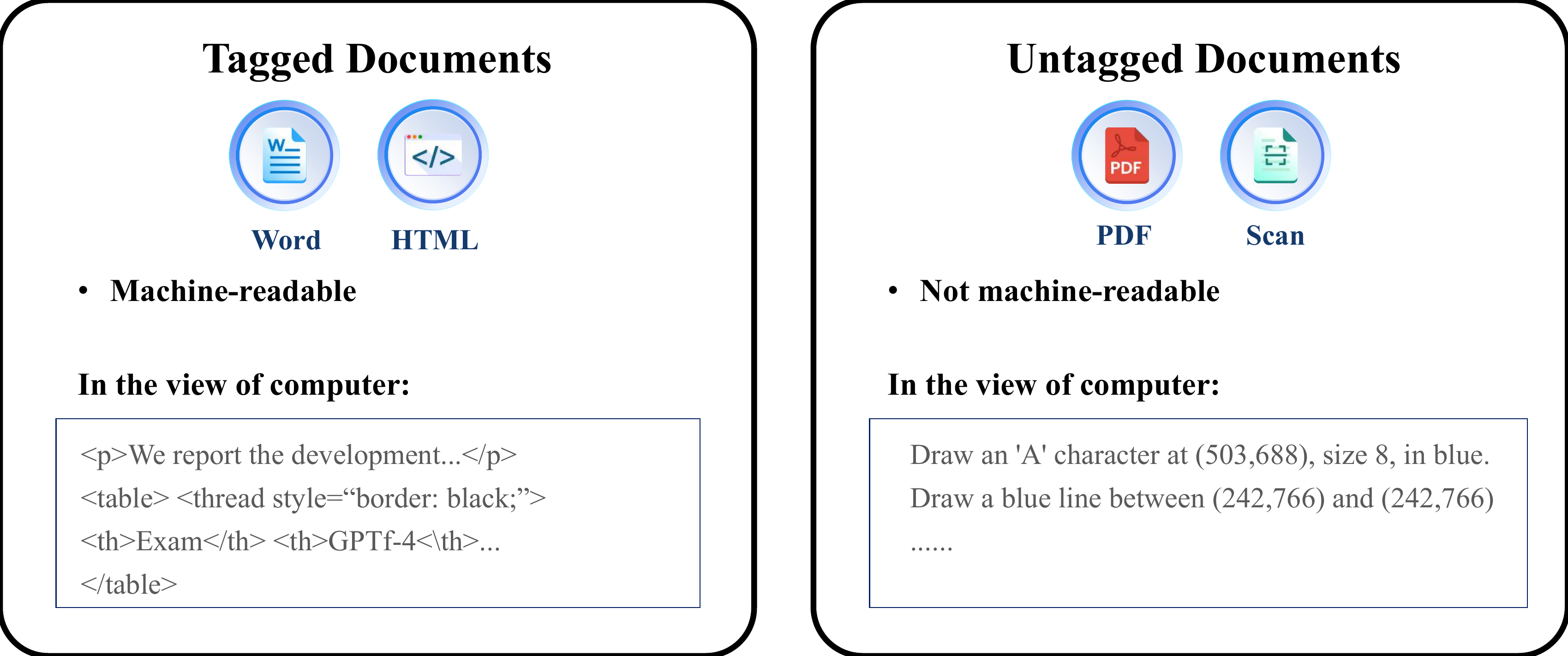}
\caption{Two types of documents in the view of computers. }
\label{fig:tagdoc}
\end{figure}

To humans, the cognitive process of perusing any document page is similar. When we read a page, characters are captured by our retinas. Then, in our brains, these characters are organized into paragraphs, tables, and charts, and then understood or memorized. However, computers perceive information as binary codes. From their perspective, as illustrated in \ref{fig:tagdoc}, documents can be categorized into two distinct types:

\begin{itemize}[leftmargin=0.5cm]
  \item Tagged Documents: Examples include Microsoft Word and HTML documents, which contain special tags like \textless p\textgreater  and \textless table\textgreater \ to organize the text into paragraphs, cells, and tables. 
  
  \item Untagged Documents: Examples include PDFs, which store instructions on the placement of characters, lines, and other content elements on each document page. They focus on 'drawing' these basic content elements in a way that makes the document legible to human readers. They do not store any structural information of the document, like tables or paragraphs. Thus, untagged documents are only for human e-reading, but are unreadable by machines. This becomes evident when attempting to copy a table from a PDF into MS Word, where the original structure of the table is often completely lost. 
  
\end{itemize}

However, Large Language Models (LLMs) exhibit proficiency in processing serialized text. Consequently, to enable LLMs to effectively manage untagged documents, a parser that organizes scattered characters into coherent texts with their structures is necessary. Ideally, a PDF Parser should exhibit the following key features:

\begin{itemize}[leftmargin=0.5cm]
  \item \textbf{Document Structure Recognition}: It should adeptly divide pages into different types of content blocks like paragraphs, tables, and charts. This ensures that the divided text blocks are complete and independent semantic units. 
  \item \textbf{Robustness in Complex Document Layout}: It should work well even for document pages with complex layouts, such as multi-column pages, border-less tables, and even tables with merged cells.
\end{itemize}

Currently, there are two main types of methods of PDF Parsing: rule based approaches and deep learning-based approaches. Among them, PyPDF, a widely-used rule-based parser, is a standard method in LangChain for PDF parsing. Conversely, our approach, ChatDOC PDF Parser (\url{https://pdfparser.io/}), is grounded in the deep learning models. Next, we illustrate the difference between them by introducing the methods and delving into some real-world cases.

\subsection{Rule-based Method: PyPDF}

We first introduce the parsing \& chunking workflow based on PyPDF. First, PyPDF serializes characters in a PDF into a long sequence without document structure information. Then, this sequence undergoes segmentation into discrete chunks, utilizing some segmentation rule, such as the ``\texttt{RecursiveCharacterTextSplitter}'' function in LangChain. Specifically, this function divides the document based on a predefined list of separators, such as the newline character ``\textbackslash n''. After this initial segmentation, adjacent chunks are merged only if the length of the combined chunks is not bigger than a predetermined limit of $N$ characters. Hereafter, we use ``PyPDF'' to refer to the method of document parsing and chunking using PyPDF+\texttt{RecursiveCharacterTextSplitter}, provided there is no contextual ambiguity. The maximum length of a chunk is set to 300 tokens in the following. Next, we use a case to observe the inherent nature of PyPDF.

\begin{figure}[t]
\centering
\includegraphics[width=1\textwidth]{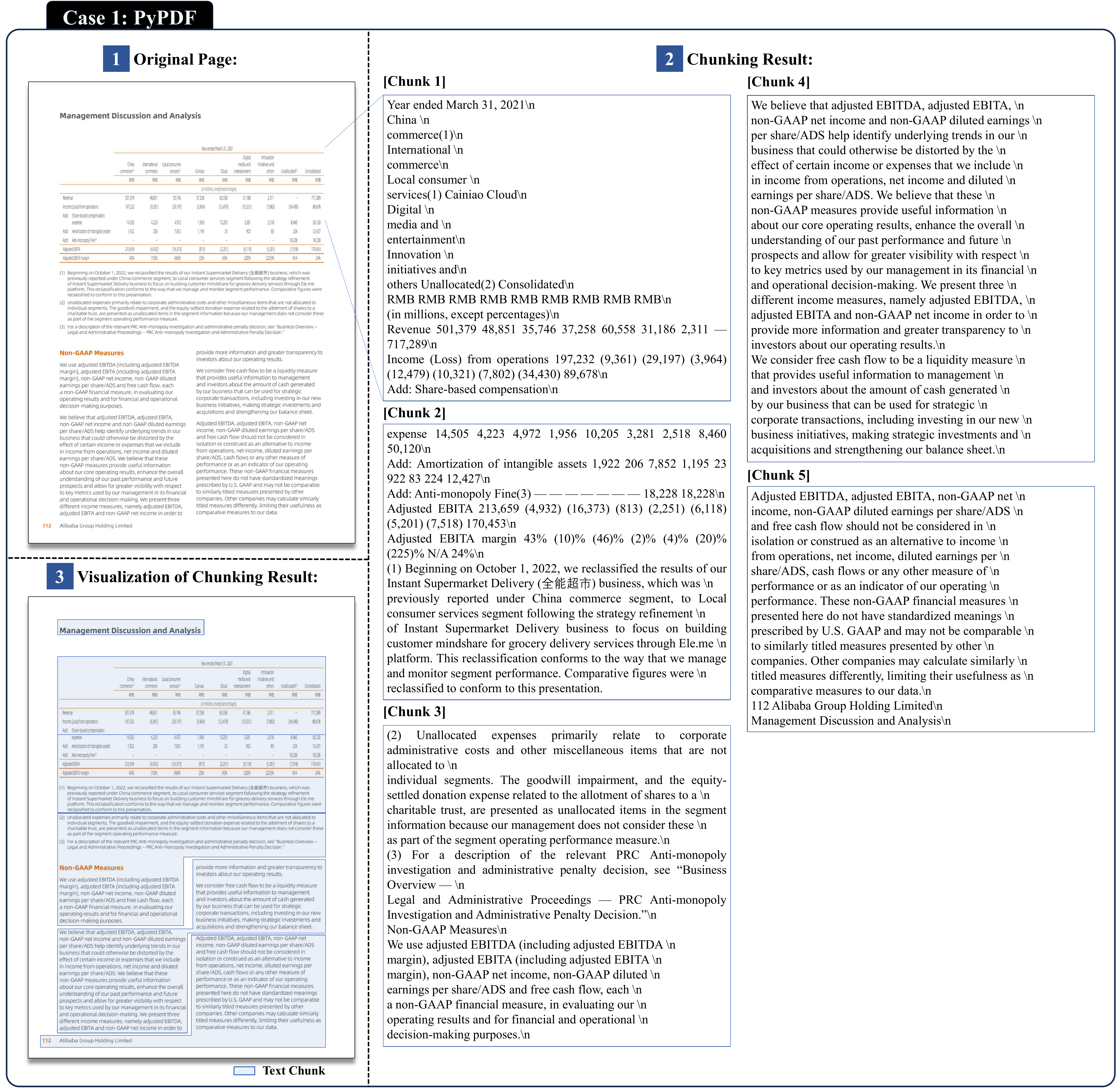}
\caption{Parsing and chunking results of PyPDF on \nameref{par:case1} (original document: \cite{alibaba}). Zoom in to see the details.}
\label{fig:case2pypdfnew}
\end{figure}

\paragraph{Case 1}\label{par:case1} in~\ref{fig:case2pypdfnew} is a page from a document that features a mix of a table and double-column text where their boundaries are difficult to distinguish. Rows in the middle of the table do not have horizontal lines, making it difficult to recognize the rows in the table. And paragraphs have both single-column layout (for notes below the table) and double-columns layout (for paragraphs in the lower part of the page).

The parsing and chunking result of PyPDF is shown in \ref{fig:case2pypdfnew}. In the ``3 Visualization'' part, we can see that PyPDF correctly recognizes the one-column and two-column layout parts of the page. But there are three shortcomings of PyPDF:
\begin{enumerate}[leftmargin=0.5cm]
    \item It cannot recognize the boundary of paragraphs and tables. It wrongly splits the table into two parts and merges the second part with the subsequent paragraph as one chunk. 
    
    PyPDF seems to be good at detecting the boundary of a paragraph, as it does not split one paragraph into multiple chunks. But it actually does not parse the boundary of a paragraph. In the ``2 Chunking Result'' part we can see that each visual text line in the page is parsed as a line ended with ``\textbackslash n'' in the result, and there is no special format at the end of a paragraph. It chunks paragraphs correctly because we use a special separator ``.\textbackslash n'' that regards a line ending with a period as likely to be the end of a paragraph. However, this heuristic may not hold in many cases.
    \item It cannot recognize the structure within a table. In the ``2 Chunking Result'' part, in chunk1, the upper part of the table is represented as a sequence of short phrases, where a cell may be split into multiple lines (e.g. the cell ``China commerce(1)") and some adjacent cells may be arranged in one line (e.g. the third to the fifth cells in the second line, ``services(1) Cainiao Cloud"). So, the structure of the table is completely destroyed. If this chunk is retrieved for RAG, LLM is unable to perceive any meaningful information from it. Similar situation for Chunk 2. Moreover, the headers of the table only exist in Chunk 1, so the lower part of the table in Chunk 2 becomes meaningless.
    \item It cannot recognize the reading order of the content. The last line of Chunk 5, ``Management Discussion and Analysis'' is actually located at the top of the page, but is parsed as the last sentence in the result. This is because PyPDF parses the document by the storage order of the characters, instead of their reading order. This may cause chaotic results when faced with complex layouts.
\end{enumerate}

The result on another case \nameref{par:case2} features with a complex and cross-page table is shown in \ref{fig:case1pypdf} in the Appendix.

\subsection{Deep Learning-based Method: ChatDOC PDF Parser}

Next, we turn our attention to the method of deep learning-based parsing, exemplified by our ChatDOC PDF Parser. The ChatDOC PDF Parser  (\url{https://pdfparser.io/}) has been trained on a corpus of over ten million document pages. Following the method in~\cite{DBLP:conf/icdar/Cao0Z021}, it incorporates a sequence of sophisticated steps, including:

\begin{enumerate}[leftmargin=1cm, parsep=0.1em]
  \item OCR for text positioning and recognition;
  \item Physical document object detection;
  \item Cross-column and cross-page trimming;
  \item Reading order determination;
  \item Table structure recognition;
  \item Document logical structure recognition.
\end{enumerate}
Readers might refer to~\cite{DBLP:conf/icdar/Cao0Z021} for the details of these steps. After parsing, we use the paragraphs and tables as basic blocks, and merge adjacent blocks until reaching the token limit to form a chunk.

\begin{figure}[t]
\centering
\includegraphics[width=1\textwidth]{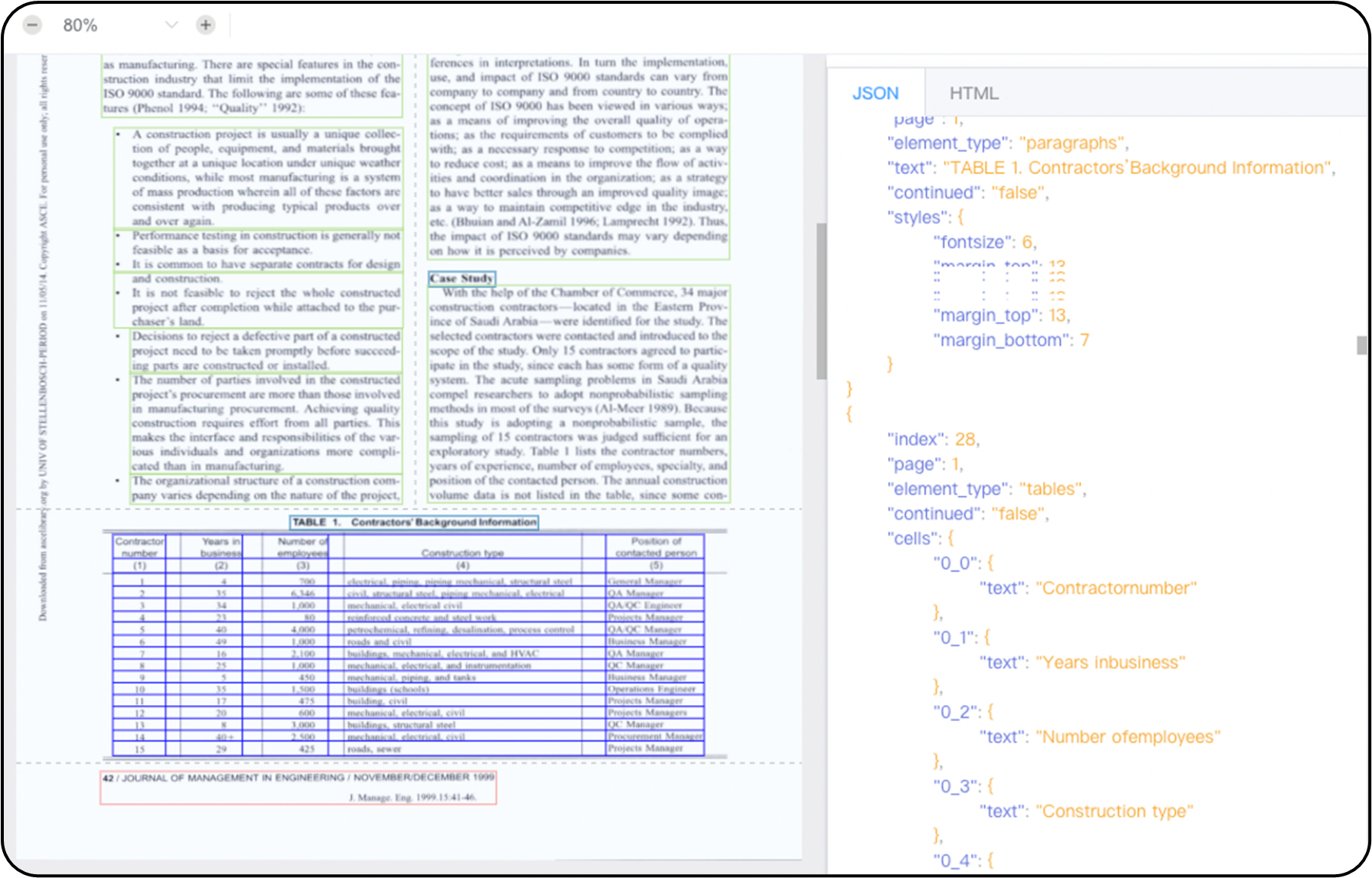}
\caption{An example illustrating the results of the ChatDOC PDF Parser. Zoom in to see the details.}
\label{fig:parserexample}
\end{figure}

ChatDOC PDF Parser is designed to consistently deliver parsing results in JSON or HTML formats, even for challenging PDF files. It parses a document into content blocks where each block refers to a table, paragraph, chart, or other type. For tables, it outputs the text in each table cell and also tells which cells are merged into a new one.  Moreover, for documents with hierarchical headings, it outputs the hierarchical structure of the document. In summary, the parsed result is like a well-organized Word file. \ref{fig:parserexample} shows a scan-copy page and its parsing result. The left side displays the document and the recognized content blocks (with different colored rectangles). The right side shows the parsing result in JSON or HTML format. Readers might refer to~\cite{pdfparser} for the live demo of this parsing result.

\begin{figure}[t]
\centering
\includegraphics[width=1\textwidth]{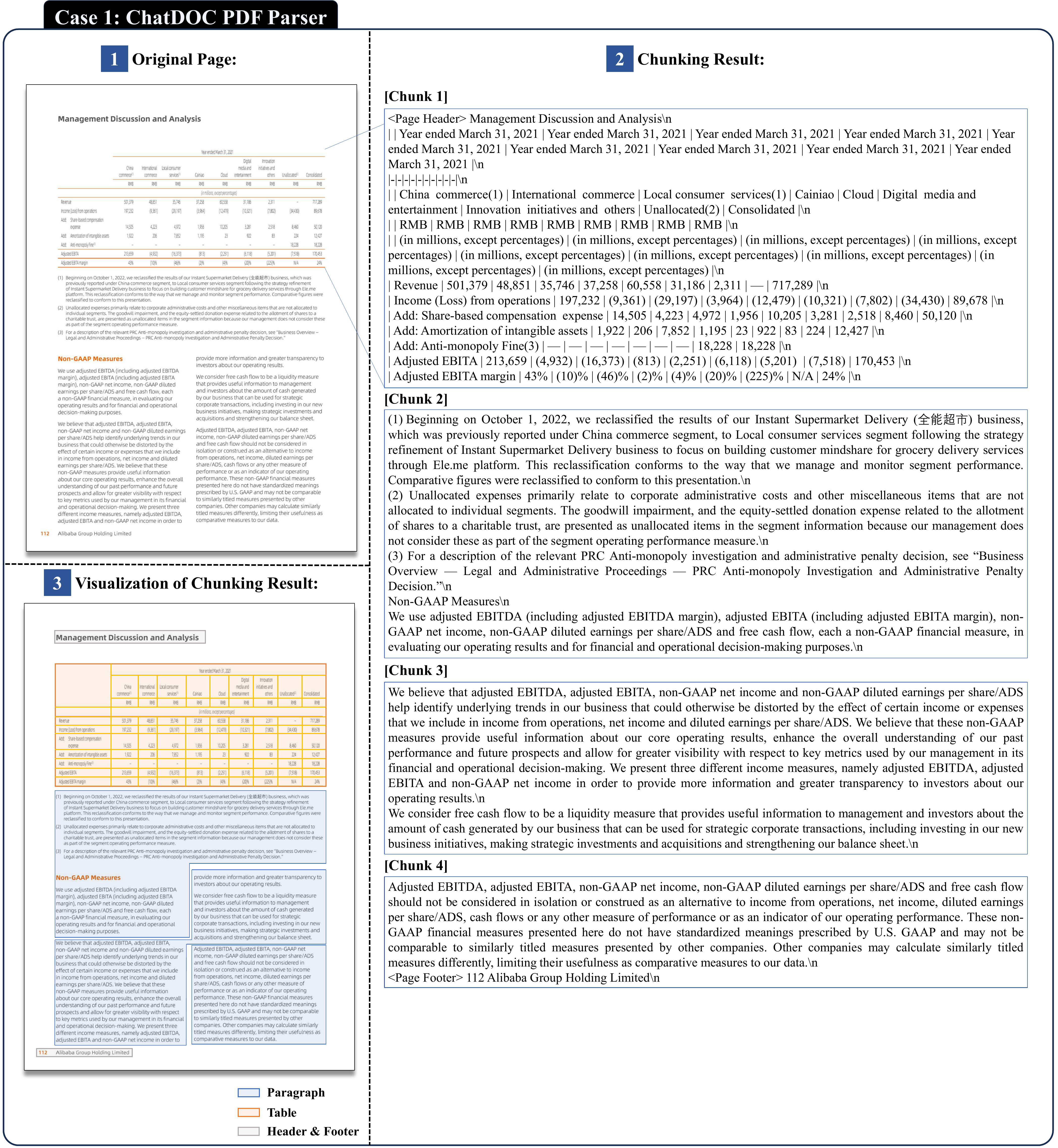}
\caption{Parsing and chunking results of ChatDOC PDF Parser on \nameref{par:case1}  (original document: \cite{casedaisho}). Zoom in to see the details.}
\label{fig:case2chatdocnew}
\end{figure}

Then, we check the result of ChatDOC PDF Parser on \nameref{par:case1} in \ref{fig:case2chatdocnew}. It successfully addresses the three shortcomings of PyPDF. 
\begin{enumerate}[leftmargin=0.5cm]
    \item As shown in the ``3 Visualization'' part, it recognizes the mixed layout and correctly sets the whole table as a separate chunk. For paragraphs, as shown in chunk 2 in the  ``2 Chunking Result'' part, text lines in the same paragraphs are merged together, making it easier to understand. 
    \item In the ``2 Chunking Result'' part, in Chunk 1, we can see the table is represented using the \texttt{markdown} format, which preserves the table's internal structure. Additionally, ChatDOC PDF Parser can recognize the merged cells inside a table. Since the \texttt{markdown} format cannot represent the merged cells, we put the whole text in the merged cell into each original cell in the \texttt{markdown} format. As you can see, in Chunk 1 the text ``Year ended March 31, 2021" repeats 9 times, which stands for a merged cell with the original 9 ones. 
    \item Moreover, ``Management Discussion and Analysis'' and ``112 Alibaba Group Holding Limited'' is recognized as the page header and footer, and they are placed at the top and bottom of the parsing result which is consistent with reading order.
\end{enumerate}

The result on another case of \nameref{par:case2} featured with complex and cross-page table is shown in \ref{fig:case1chatdoc} in the Appendix.

\section{Experiments on the Impact of PDF Recognition on RAG}

Back to the main topic of this paper, does the way a document is parsed and chunked affect the quality of answers provided by an RAG system? To answer this, we have carried out a systematic experiment to assess the impacts.

\subsection{Quantitative Evaluation of RAG Answer Quality}

\subsubsection{Settings}

We compared two RAG systems as listed in \ref{tab:ragoverview}: 

\begin{itemize}[leftmargin=0.5cm]
    \item ChatDOC: uses ChatDOC PDF Parser to parse the document and leverage the structure information for chunking.
    \item Baseline: uses PyPDF to parse the document and use \texttt{RecursiveCharacterTextSplitter} function for chunking.
\end{itemize}

Other components, like embedding, retrieval, and QA, are the same for both systems. 

\begin{table}[t]
    \centering
    \begin{tabular}{ccc}
    \specialrule{0.09em}{3pt}{3pt}
        \textbf{Steps $\downarrow$} & \begin{tabular}[c]{@{}c@{}}\textbf{ChatDOC} \\{\footnotesize (PDFlux-LLM)}\end{tabular} & \begin{tabular}[c]{@{}c@{}}\textbf{Baseline} \\{\footnotesize (PyPDF-LLM)}\end{tabular} \\ \specialrule{0.05em}{3pt}{3pt}
        \textbf{PDF Parsing} & \begin{tabular}[c]{@{}c@{}}\textbf{PDFlux} \\{\footnotesize (Deep Learning-based)}\end{tabular} & \begin{tabular}[c]{@{}c@{}}\textbf{PyPDF} \\{\footnotesize (Rule-based, default method in LangChain)}\end{tabular} \\ \specialrule{0em}{3pt}{3pt}
        \textbf{Chunking} & \begin{tabular}{@{}c@{}} $\approx$300 tokens per chunk \\ + chunking via paragraphs, tables etc.\end{tabular} & $\approx$300 tokens per chunk + separator \\ \specialrule{0em}{3pt}{3pt}
        \textbf{Embedding} & \multicolumn{2}{c}{text-embedding-ada-002} \\ \specialrule{0em}{3pt}{3pt}
        \textbf{Retrieval} & \multicolumn{2}{c}{$\le$3000 tokens} \\ \specialrule{0em}{3pt}{3pt}
        \textbf{QA} & \multicolumn{2}{c}{GPT3.5-Turbo} \\ \specialrule{0.09em}{3pt}{3pt}
    \end{tabular}
    \caption{Settings of two RAG systems: ChatDOC and Baseline.}
    \label{tab:ragoverview}
\end{table}

\subsubsection{Data Preparation}

For our experiment, we assembled a dataset that closely mirrors real-world conditions, comprising 188 documents from various fields. Specifically, This collection includes 100 academic papers, 28 financial reports, and 60 documents from other categories such as textbooks, courseware, and legislative materials.

\begin{table}[t]
    \centering
    \begin{tabular}{m{1.5cm}m{5cm}m{5.5cm}}
    \specialrule{0.09em}{3pt}{3pt}
        ~ & \textbf{Extractive Questions} & \textbf{Comprehensive Analysis Questions} \\ \specialrule{0.05em}{3pt}{3pt}
        \textbf{Number} & 86 & 216 \\ 
       \specialrule{0.05em}{3pt}{3pt}
        \textbf{Question Examples} & 
            \textit{1. Locate the content of section ten, what is the merged operating cost in the income statement? 
            \newline 2. What is the specific content of table 1.
            \newline 3. Extract financial data and profit forecast tables.
            \newline 4. Find the long-term loan table.}
            &
            \textit{1. Summarize and analyze the profit forecast and valuation in the research report.
            \newline 2. Fully report the research approach of this text.
            \newline 3. Analyze the long-term debt-paying ability based on this report.
            \newline 4. How is the feasibility analysis done in this article?
            \newline 5. Give a simple example to explain the encoding steps and algorithm in the paper.}
            \\ \specialrule{0.05em}{3pt}{3pt}
        \textbf{Evaluation} & Human Evaluation & GPT 4 evaluation \\ \specialrule{0.09em}{3pt}{3pt}
    \end{tabular}
    \caption{The questions in the dataset are categorized into extractive questions and comprehensive analysis questions. }
    \label{tab:questionoverview}
\end{table}

We then gathered 800 manually generated questions via crowd-sourcing. After careful screening, we removed low-quality questions and got 302 questions for evaluation. These questions were divided into two categories (as shown in \ref{tab:questionoverview}):
\begin{itemize}[leftmargin=0.5cm]
  \item \textbf{Extractive questions} are those that can be answered with direct excerpts from the documents. Usually, they require pinpoint answers because they seek specific information. We found when using LLM for evaluation, it may fail to distinguish the subtle but important differences between answers, so we relied on human assessment. We used a 0-10 scale to rate the results. An annotator is given the retrieved content and answer of both methods and rates the two methods at the same time. We show the retrieved content because it usually cannot evaluate the answer without document content, and show two methods together to promote detailed comparison, especially on partially correct results.
  \item \textbf{Comprehensive analysis questions} necessitate synthesizing information from multiple sources and aspects and making a summary. Since the answer is lengthy and requires a comprehensive understanding of the given document contents, we found it difficult and time-consuming for humans to evaluate. Hence, we used GPT-4 to evaluate 
 the answer quality, scoring from 1-10. We also rate the result of two methods in one request. But we only give the retrieved content without an answer because the answer is lengthy (thus costly) compared with extractive questions and a better retrieved content can imply a better answer (since the used LLM is the same).  A pair of results of two methods is scored 4 times to avoid bias~\cite{wang2023large}, and their average value is used. Specifically, for a pair of content ($A$, $B$) to be compared for the same question, we feed both $A$ and $B$ to GPT-4 to compare and score them twice. We also flip their order, feed $B$ and $A$ to GPT-4, and repeat the request twice.
\end{itemize}

\subsubsection{Results}

\textbf{Results of Extractive Questions}

The results of extractive questions are shown in \ref{tab:experimentres}. Out of the 86 extractive questions, ChatDOC performed better than the baseline on 42 cases, tied on 36 cases, and was inferior to Baseline on only 8 cases.

\begin{table}[t]
    \centering
    \begin{tabular}{m{2.7cm}m{1cm}m{2.7cm}m{1.5cm}m{2.7cm}}
    \specialrule{0.09em}{3pt}{3pt}
        ~ & \centering \textbf{Total} & \centering\arraybackslash \textbf{ChatDOC wins} & \centering\arraybackslash \textbf{Tie} & \centering\arraybackslash \textbf{Baseline wins} 
        \\ \specialrule{0.05em}{3pt}{3pt}
        \centering\arraybackslash \textbf{Extractive Questions} & 
        \centering 86 & 
        \parbox[b]{2.7cm}{\centering \textbf{42} \\ [2pt]{\footnotesize \textbf{(49\%)}} } & 
        \parbox[b]{1.5cm}{\centering 36 \\ [2pt]{\footnotesize (42\%)} } & 
        \parbox[b]{2.7cm}{\centering 8 \\ [2pt]{\footnotesize (9\%)} } 
        \\ \specialrule{0em}{3pt}{3pt}
        \centering\arraybackslash \textbf{Comprehensive Questions} & 
        \centering 216 & 
        \parbox[b]{2.7cm}{\centering \textbf{101} \\ [2pt]{\footnotesize \textbf{(47\%)}} } & 
        \parbox[b]{1.5cm}{\centering 79 \\ [2pt]{\footnotesize (37\%)} } & 
        \parbox[b]{2.7cm}{\centering 36 \\ [2pt]{\footnotesize (17\%)} } 
        \\ \specialrule{0.05em}{3pt}{3pt}
        \centering\arraybackslash \textbf{Summary} & 
        \centering 302 & 
        \parbox[b]{2.7cm}{\centering \textbf{143} \\ [2pt]{\footnotesize \textbf{(47\%)}} } & 
        \parbox[b]{1.5cm}{\centering 115 \\ [2pt]{\footnotesize (38\%)} } & 
        \parbox[b]{2.7cm}{\centering 44 \\ [2pt]{\footnotesize (15\%)} }
        \\ \specialrule{0.09em}{3pt}{3pt}
    \end{tabular}
    \caption{The comparison result between ChatDOC and Baseline. }
    \label{tab:experimentres}
\end{table}

\begin{figure}[t]
\centering
\includegraphics[width=.8\textwidth]{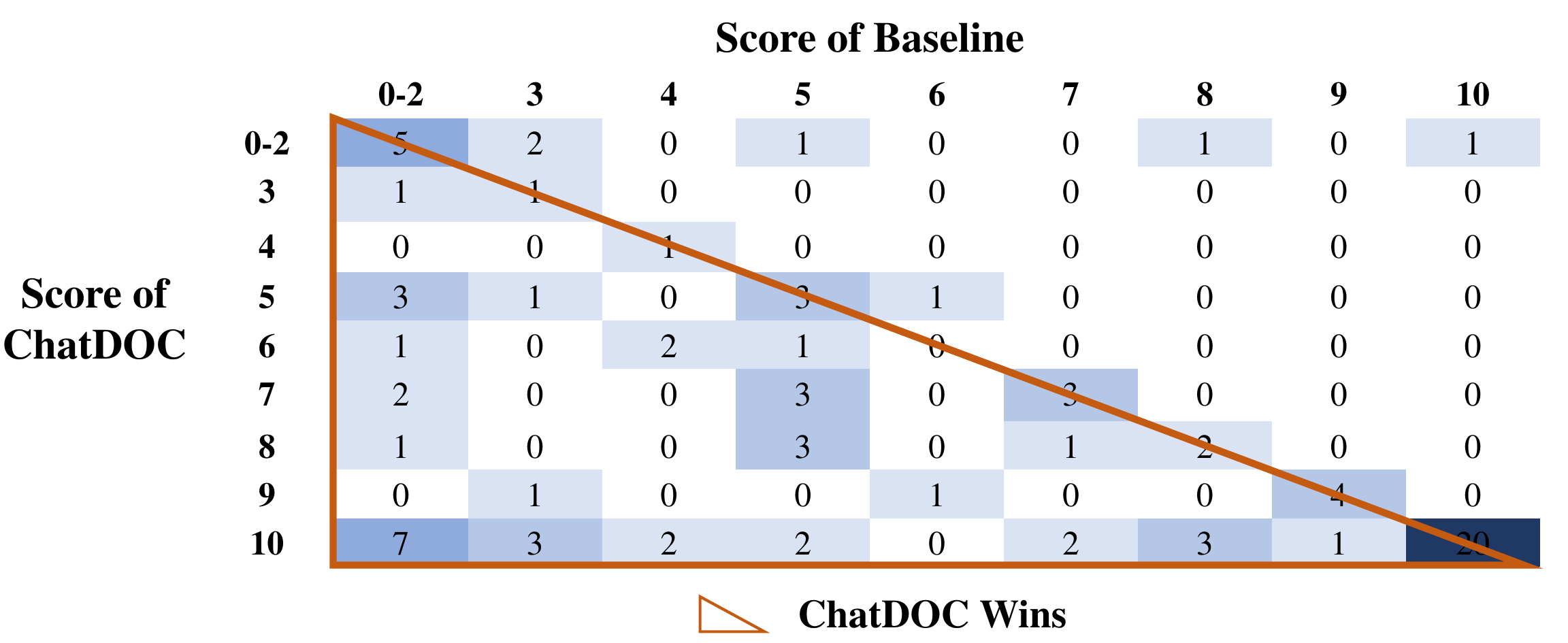}
\caption{Distribution of rating scores of extractive questions.}
\label{fig:extractivescoretable}
\end{figure}

\begin{figure}[h!]
\centering
\includegraphics[width=.8\textwidth]{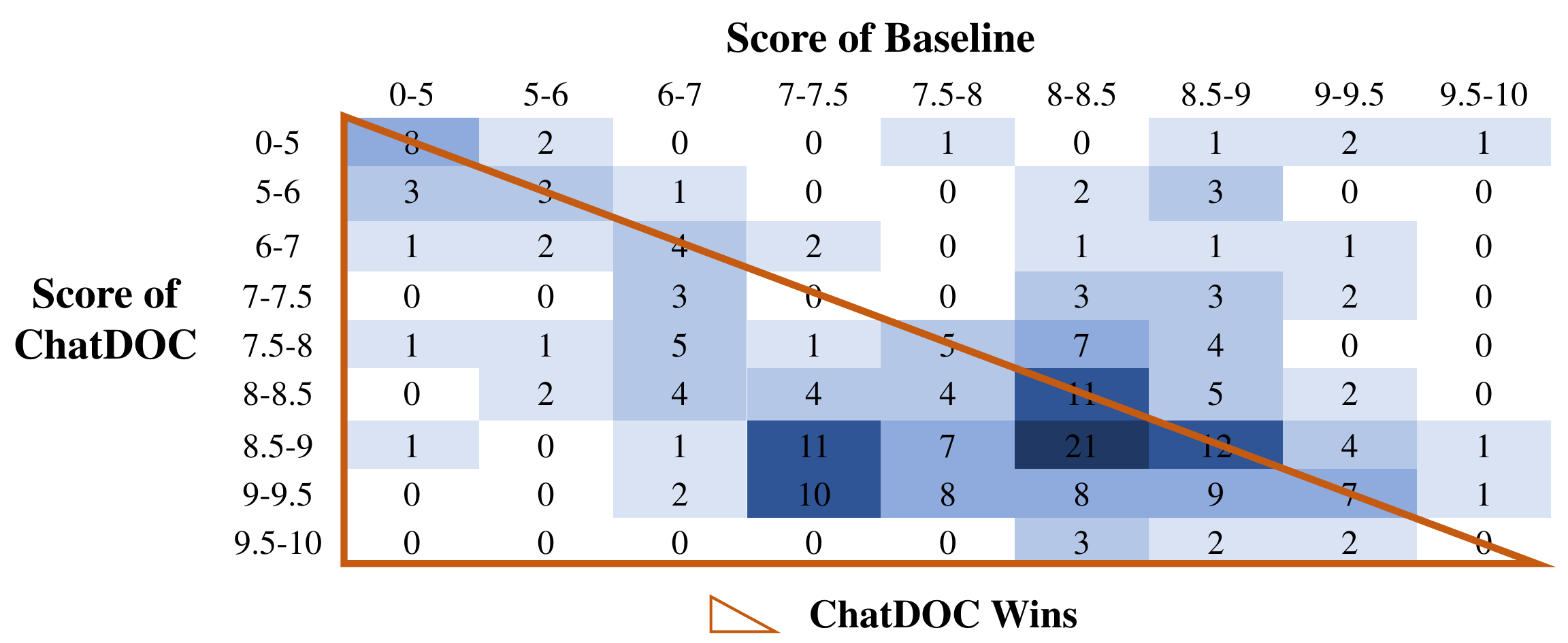}
\caption{Distribution of rating scores of comprehensive analysis questions. }
\label{fig:comprehenscoretable}
\end{figure}

The distribution of rating scores is further detailed in \ref{fig:extractivescoretable}. In the distribution table, $T_{ij}=k$ means there are $k$ questions whose answer by ChatDOC is rated as $i$ and the answer by Baseline is rated as $j$. Cases where ChatDOC scores higher than the baseline (ChatDOC wins) are represented in the lower-left half, while cases where the baseline scores higher are in the upper-right. Notably, most samples with a clear winner are in the lower-left half, indicating ChatDOC's superiority. Impressively, ChatDOC achieved full marks (10) in nearly half of these cases, amounting to a total of 40.

\textbf{Results of Comprehensive Analysis Questions}

The results of comprehensive analysis questions are shown in \ref{tab:experimentres}. Out of the 216 comprehensive analysis questions, ChatDOC performed better than the baseline on 101 cases, tied on 79 cases, and was inferior to Baseline on only 36 cases.

In \ref{fig:comprehenscoretable}, the score distribution chart for these questions shows a denser concentration in the lower-left half. This indicates that ChatDOC frequently outperforms the baseline. Notably, the majority of ChatDOC's retrievals scored between 8.5 and 9.5, demonstrating a high level of retrieval quality.

In summary, ChatDOC significantly outperforms the Baseline, primarily attributed to its superior PDF parsing method.

\subsection{Case Studies of RAG}

To make the comparison more tangible, we show some cases where ChatDOC demonstrates its superiority.
\setcounter{secnumdepth}{3}

\begin{figure}[t]
\centering
\includegraphics[width=.7\textwidth]{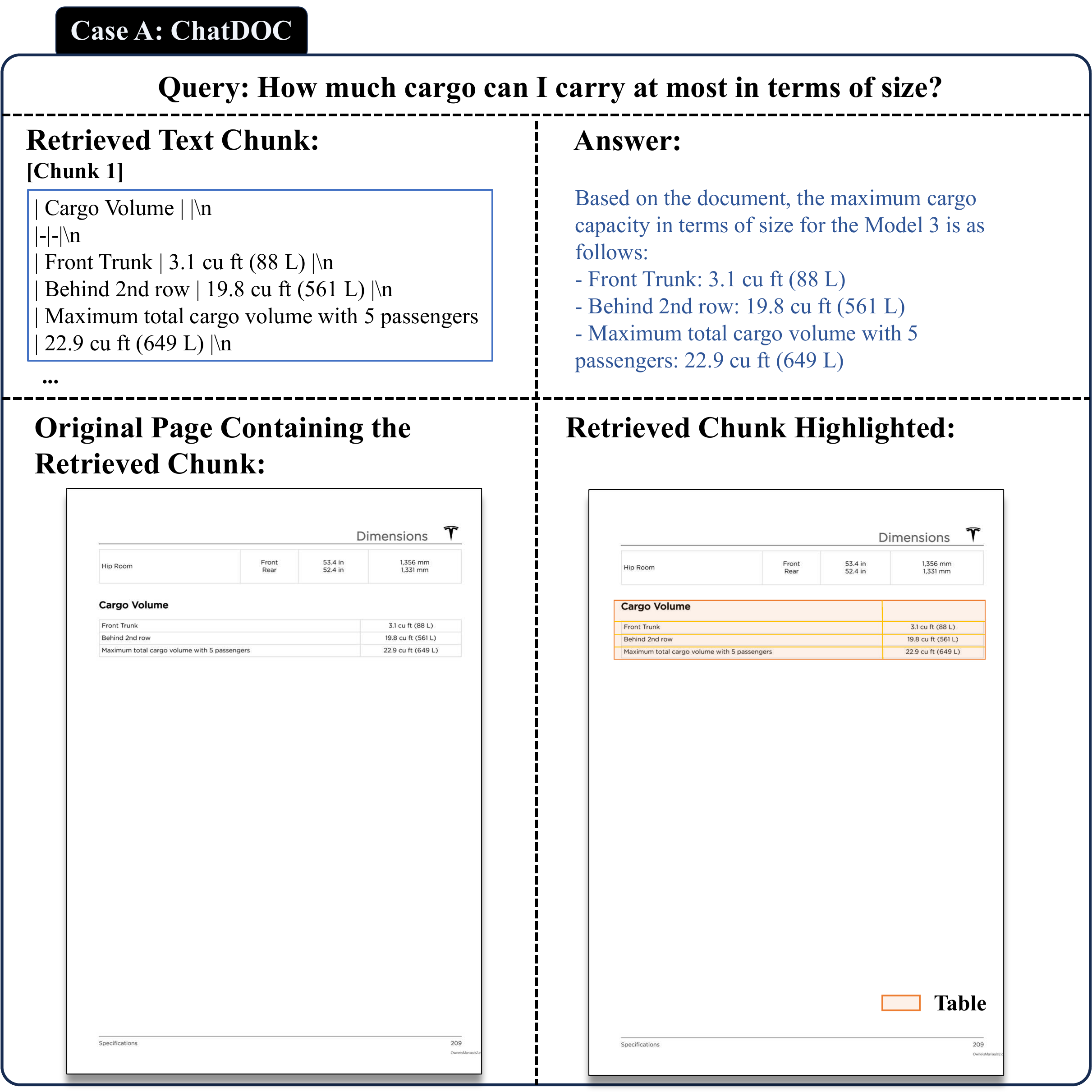}
\caption{ChatDOC's result on finding info in Tesla's Manual (original document: \cite{tesla}).}
\label{fig:expcase1chatdoc}
\end{figure}

\begin{figure}[h!]
\centering
\includegraphics[width=.7\textwidth]{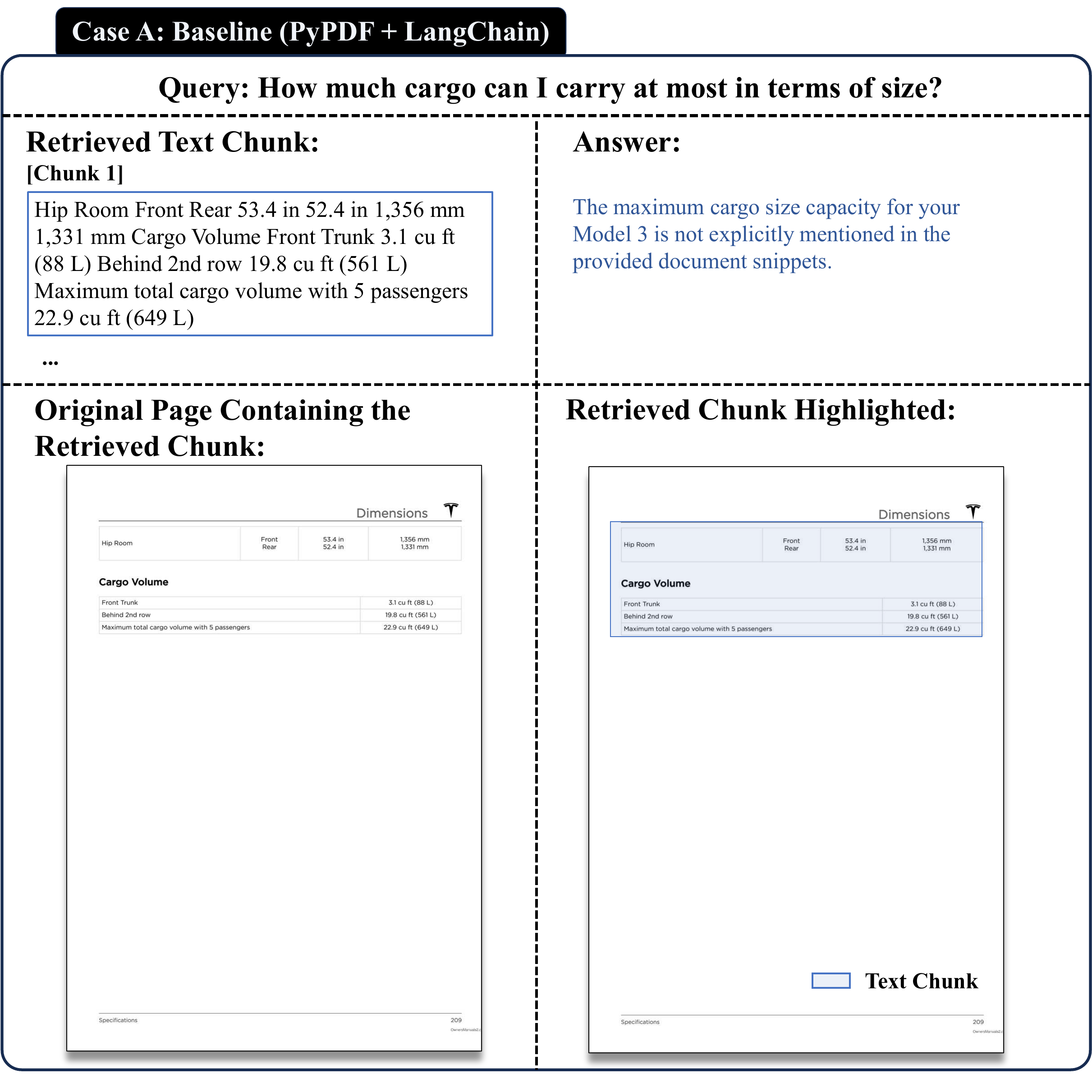}
\caption{Baseline's result on finding info in Tesla's Manual (original document: \cite{tesla}).}
\label{fig:expcase1baseline}
\end{figure}

\begin{figure}[t]
\centering
\includegraphics[width=1\textwidth]{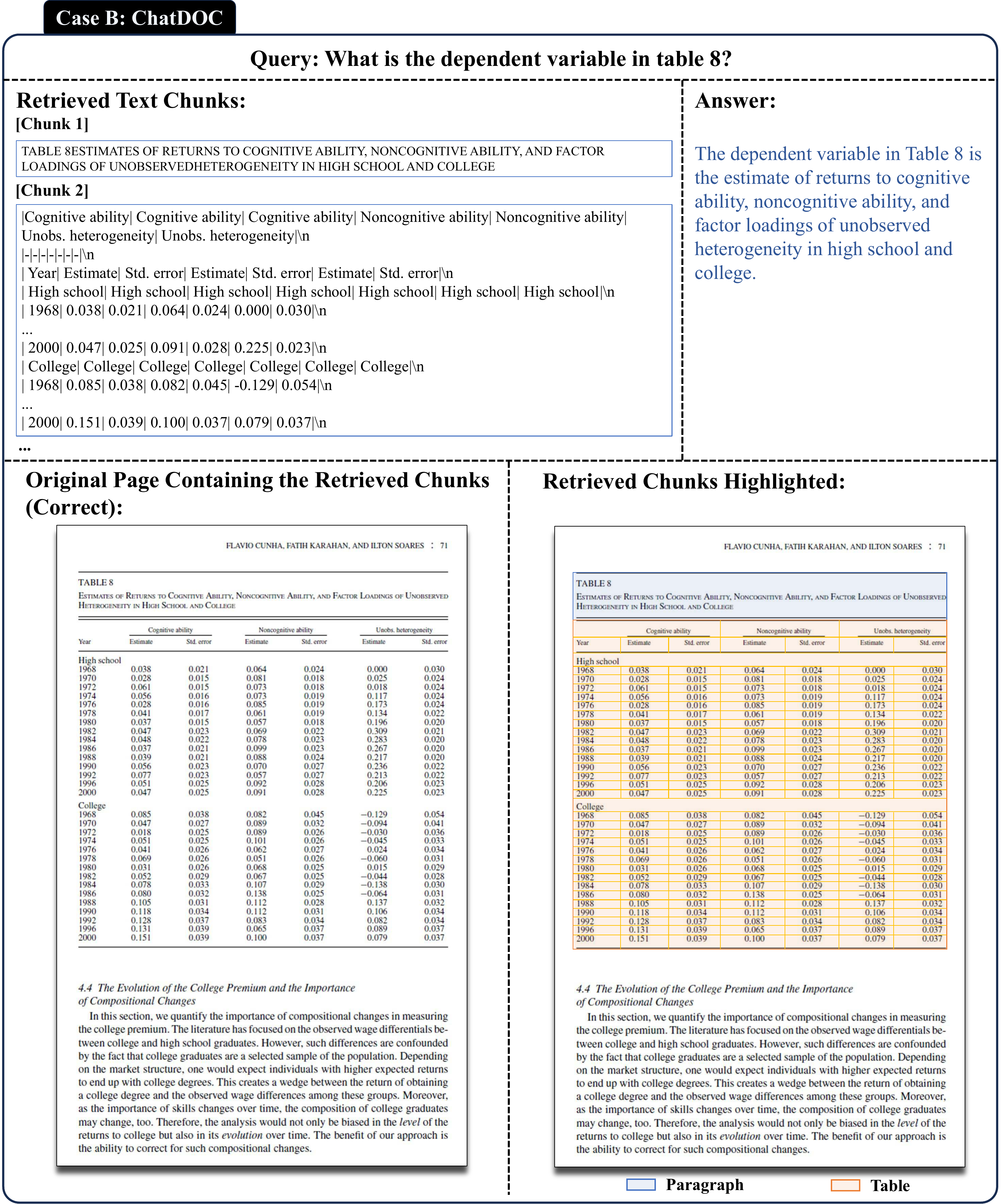}
\caption{ChatDOC's result on locating a specific table in a research paper (original document: \cite{Cunha2011ReturnsTS}).}
\label{fig:expcase2chatdoc}
\end{figure}

\subsubsection{Case A - Find Specific information in the Tesla manual}

Case A involves a query about Tesla's user manual specifically targeting cargo volume information. For this query, ChatDOC and Baseline perform differently as in \ref{fig:expcase1chatdoc} and \ref{fig:expcase1baseline}. The figures show the most relevant chunk(s) retrieved and the LLM's answer. They also show the document page that the relevant chunk(s) are located in, and highlight these chunks. In this case, both models located the table, but the text they feed to LLM is different, so the answer is different. Specifically,
\begin{itemize}[leftmargin=0.5cm]
    \item ChatDOC recognizes the table structure, interpreting the text in the \texttt{markdown} format (as shown in the ``Retrieved Text Chunks'' part), which made it easier for the language model to comprehend.
    \item Baseline erroneously merges the target table and the table above into one chunk and does not have the table structure. Hence, the text in the chunk is not understandable (as shown in the ``Retrieved Text Chunks'' part) and the LLM can only answer with ``not explicitly mentioned''.
\end{itemize}

This case underscores the effectiveness of ChatDOC's parsing method, particularly in handling tables and presenting them in an LLM-friendly format.

\subsubsection{Case B - Research paper}

\begin{figure}[t]
\centering
\includegraphics[width=1\textwidth]{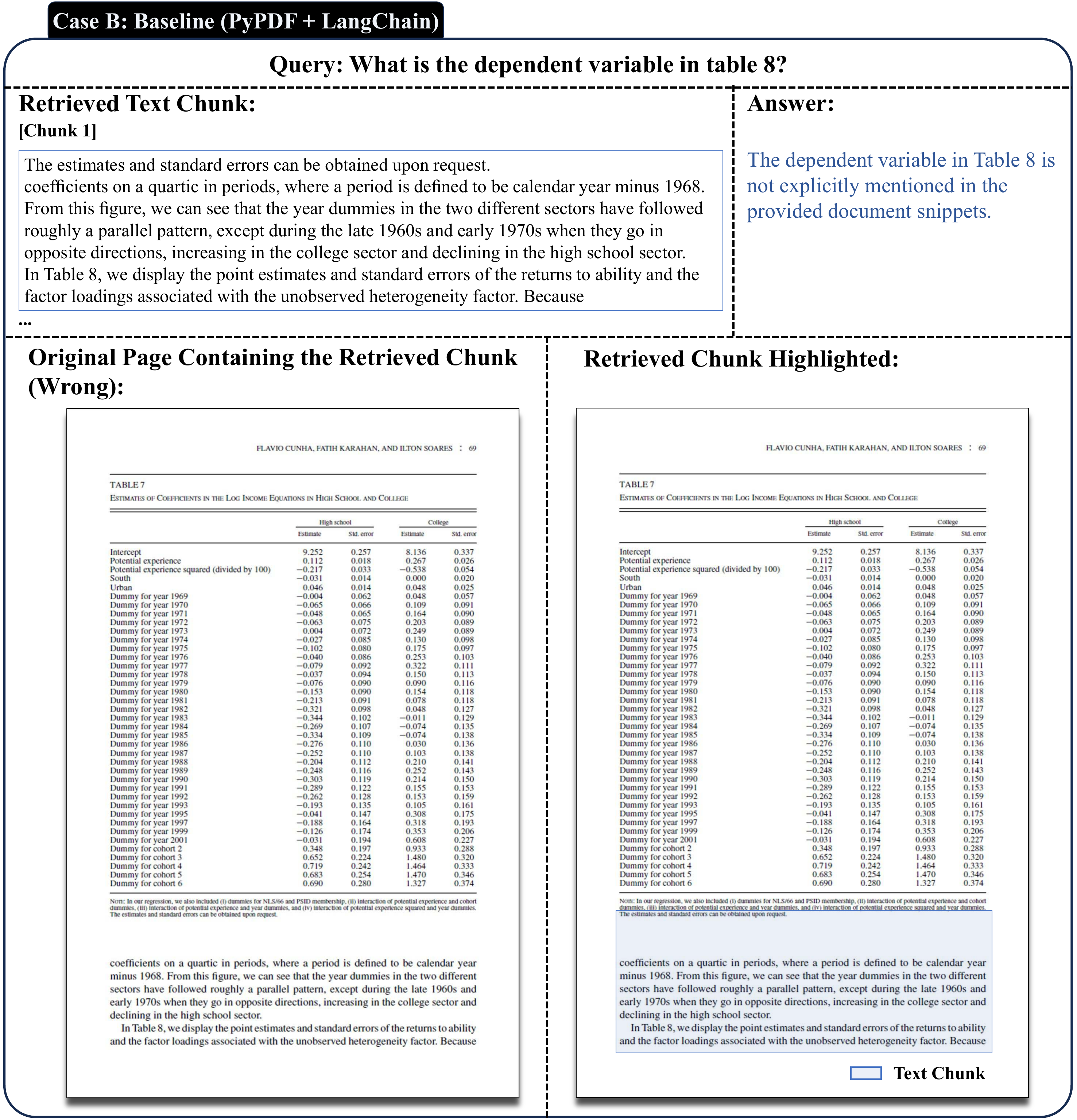}
\caption{Baseline's result in locating a specific table in a research paper (original document: \cite{Cunha2011ReturnsTS})}
\label{fig:expcase2baseline}
\end{figure}

In Case B, the user's query is on a specific research paper. It requests the system to identify ``Table 8" in the paper and enumerate all the dependent variables it lists. Both the title and the content of the table were necessary for identifying these variables. \ref{fig:expcase2chatdoc} and \ref{fig:expcase2baseline} show how ChatDOC and Baseline perform in this case.

\begin{itemize}[leftmargin=0.5cm]
    \item ChatDOC effectively retrieves the entire table, encompassing both its title and content. This comprehensive retrieval allows for an accurate response to the query.
    \item Baseline does not retrieve true ``Table 8", but only a text chunk below ``Table 7" (since it contains the text of ``Table 8). Due to the baseline’s segmentation strategy, the content of ``Table 8" and other content on the same page are combined into a large chunk. This chunk, containing a mix of unrelated content, has a low similarity score and consequently does not show up in the retrieval results.
\end{itemize}

This case highlights ChatDOC's superior ability to handle complex document structures and its impact on retrieving specific segments for accurate responses.

\begin{figure}[t]
\centering
\includegraphics[width=.85\textwidth]{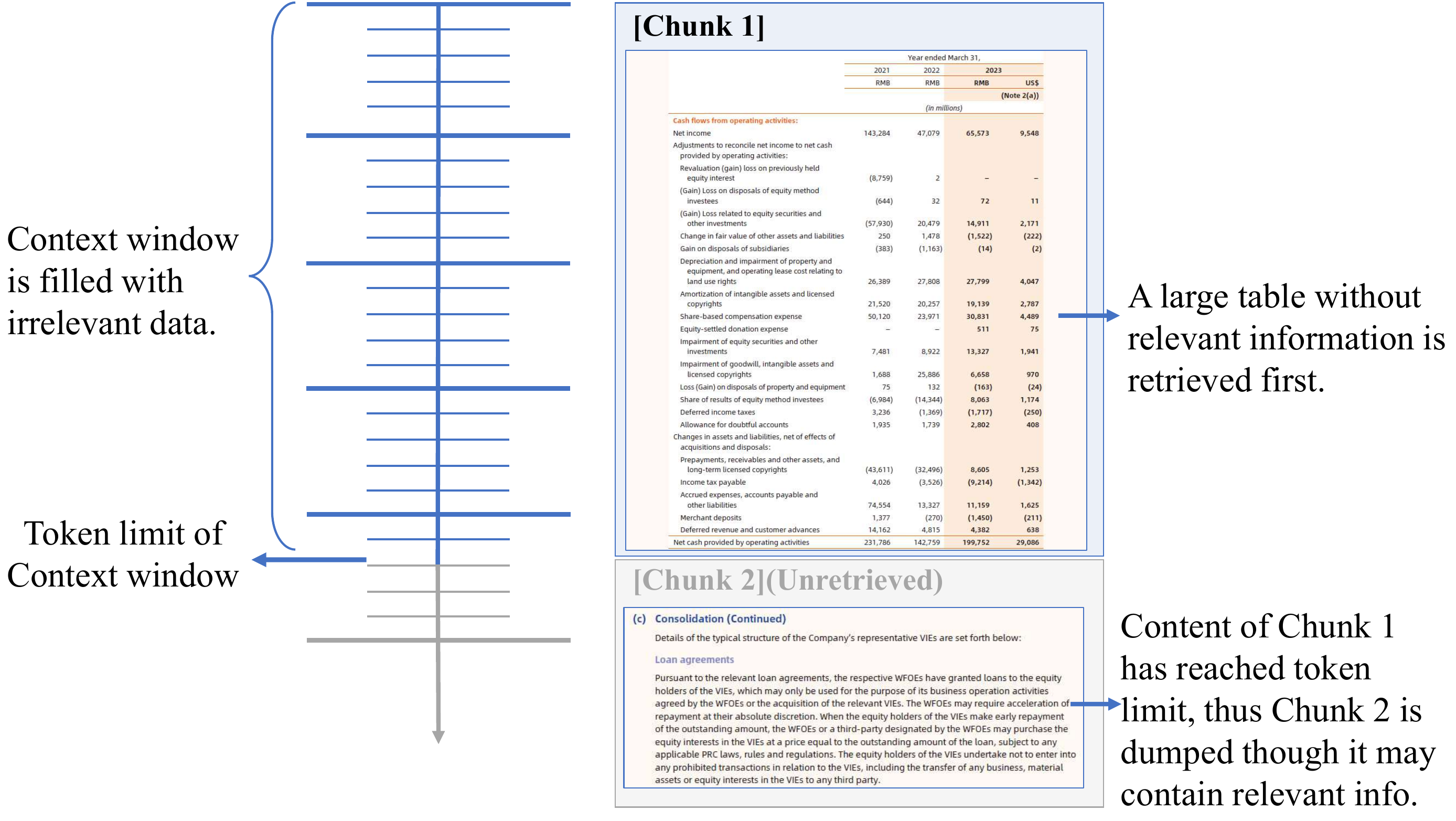}
\caption{An example of ChatDOC encountered the ranking and token limit issues.}
\label{fig:limitation1}
\end{figure}

\begin{figure}[t!]
\centering
\includegraphics[width=.85\textwidth]{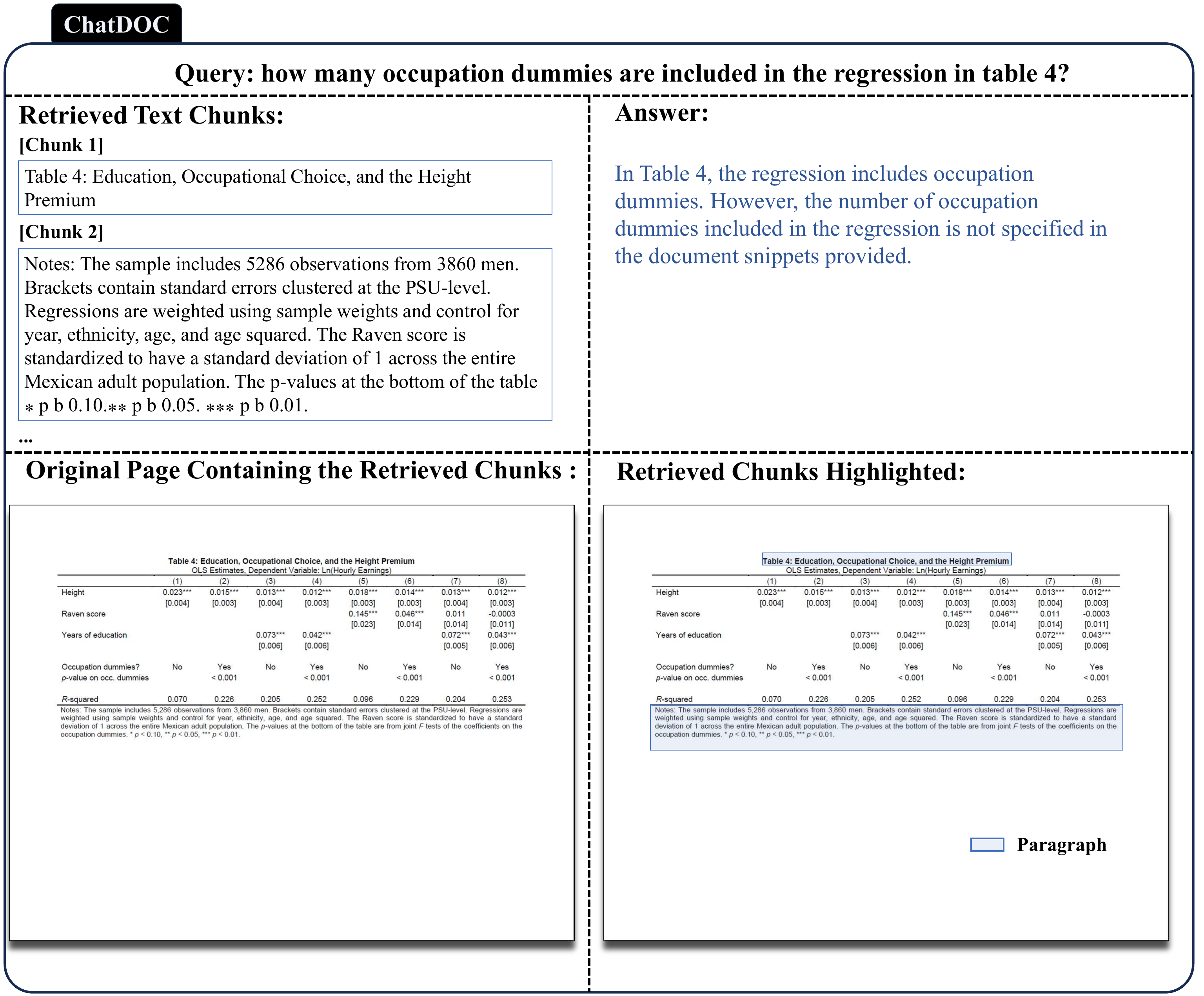}
\caption{An example that ChatDOC fails to retrieve the relevant table (original document: \cite{Vogl2012HeightSA}).}
\label{fig:limitation2}
\end{figure}

\subsection{Discussion on Limitations}
While ChatDOC generally performs well, there are instances where its retrieval quality is not as good as Baseline's. We observe two patterns in these cases.

\textbf{Ranking and Token Limit Issue.} If ChatDOC retrieves a large, but irrelevant table first, it uses up the context window, preventing access to the relevant information, as the example in \ref{fig:limitation1} shows. This is mainly because the embedding model does not rank the relevant chunk as the top result. This may be addressed by a better embedding model, or a more sophisticated way to handle large tables/paragraphs like only retaining the relevant part of the table for LLM.

\textbf{Fine Segmentation Drawback.} \ref{fig:limitation2} shows a case that requires retrieving the whole table with its title. However, ChatDOC wrongly recognizes the title as a regular paragraph, so that the title and the table are stored in different chunks. This led to retrieving only part of the required information, namely the table's title and footnotes, but not the key content within the table. Improving table title recognition could address these issues.

\section{Applications in ChatDOC}

We apply the enhanced PDF structure recognition framework on ChatDOC (\href{http://www.example.com}{chatdoc.com}), an AI file-reading assistant that helps to summarize long documents, explain complex concepts, and find key information in seconds. 

In terms of reliability and accuracy, it is the top among all ChatPDF products. Here's what makes ChatDOC special:
\begin{itemize}[leftmargin=0.5cm]
    \item Mastery over tables: Simply select any table or text, and dive right into the details.
    \item Multi-file conversation: Talk about lots of documents at the same time, without worrying about how many pages each one has.
    \item Citation-backed responses: All answers are supported by direct quotes pulled from the source documents.
    \item Handle Many File Types: Works seamlessly with scanned files, ePub, HTML, and docx formats.
\end{itemize}

We are still working on publishing the API of ChatDOC PDF Parser. Please subscribe to the wait list via  
\href{https://pdfparser.io/}{pdfparser.io}.

\section{Conclusion}

Large Language Models (LLMs) are capable of producing more accurate responses when assisted by a PDF parser that effectively extracts and integrates structured information from documents into the prompts. This process enhances the quality and relevance of the data fed into the models, thereby improving their output.

In the future, we will compare more deep learning-based document parsing methods to give a more comprehensive understanding of the relationship between the RAG quality and document parsing quality. Some initial experiments show that some open-sourced PDF parsing methods cannot meet the bar for high-quality RAG.

\vspace{1cm}

\bibliography{references}
\bibliographystyle{unsrt}

\clearpage
\appendix

\section{More Cases on PDF Parsing \& Chunking}
\begin{figure}[h!]
\centering
\includegraphics[width=1\textwidth]{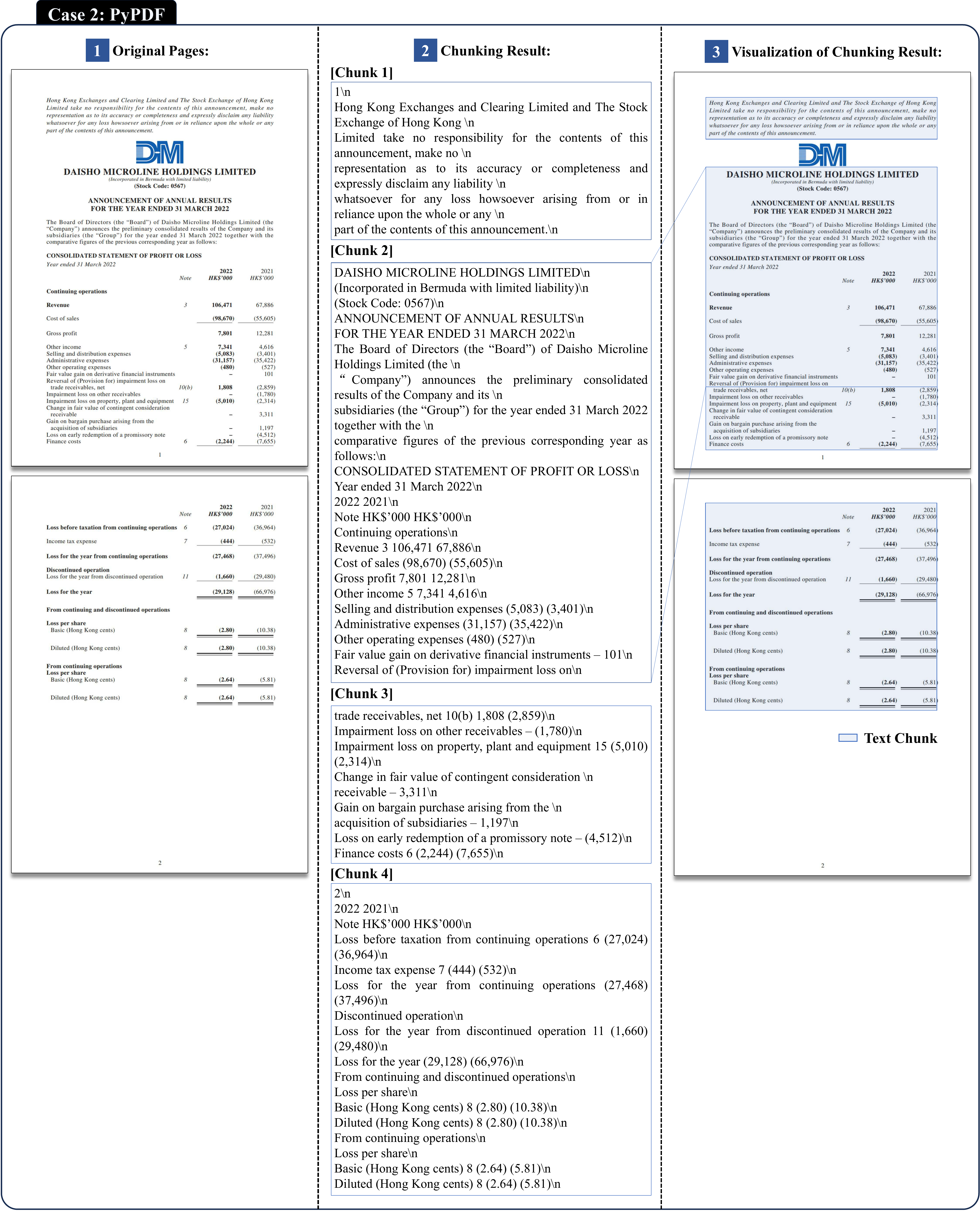}
\caption{Parsing and chunking results of PyPDF on \nameref{par:case2} (original document: \cite{casedaisho}). Zoom in to see the details.}
\label{fig:case1pypdf}
\end{figure}

\paragraph{Case 2}\label{par:case2} in \ref{fig:case1pypdf}
features a large borderless table that spans two pages. \ref{fig:case1pypdf} shows the result by PyPDF. A close inspection reveals that tables are represented merely as sequences of text, making them challenging to interpret and understand. And the table is scattered in three chunks. Results on these two cases demonstrate that the rule-based method, like that of PyPDF, tends to dissect a document without a true understanding of its content structure. As a result, tables are often torn apart and paragraphs become jumbled, leading to a disjointed and confusing representation of the original document.

\begin{figure}[h!]
\centering
\includegraphics[width=1\textwidth]{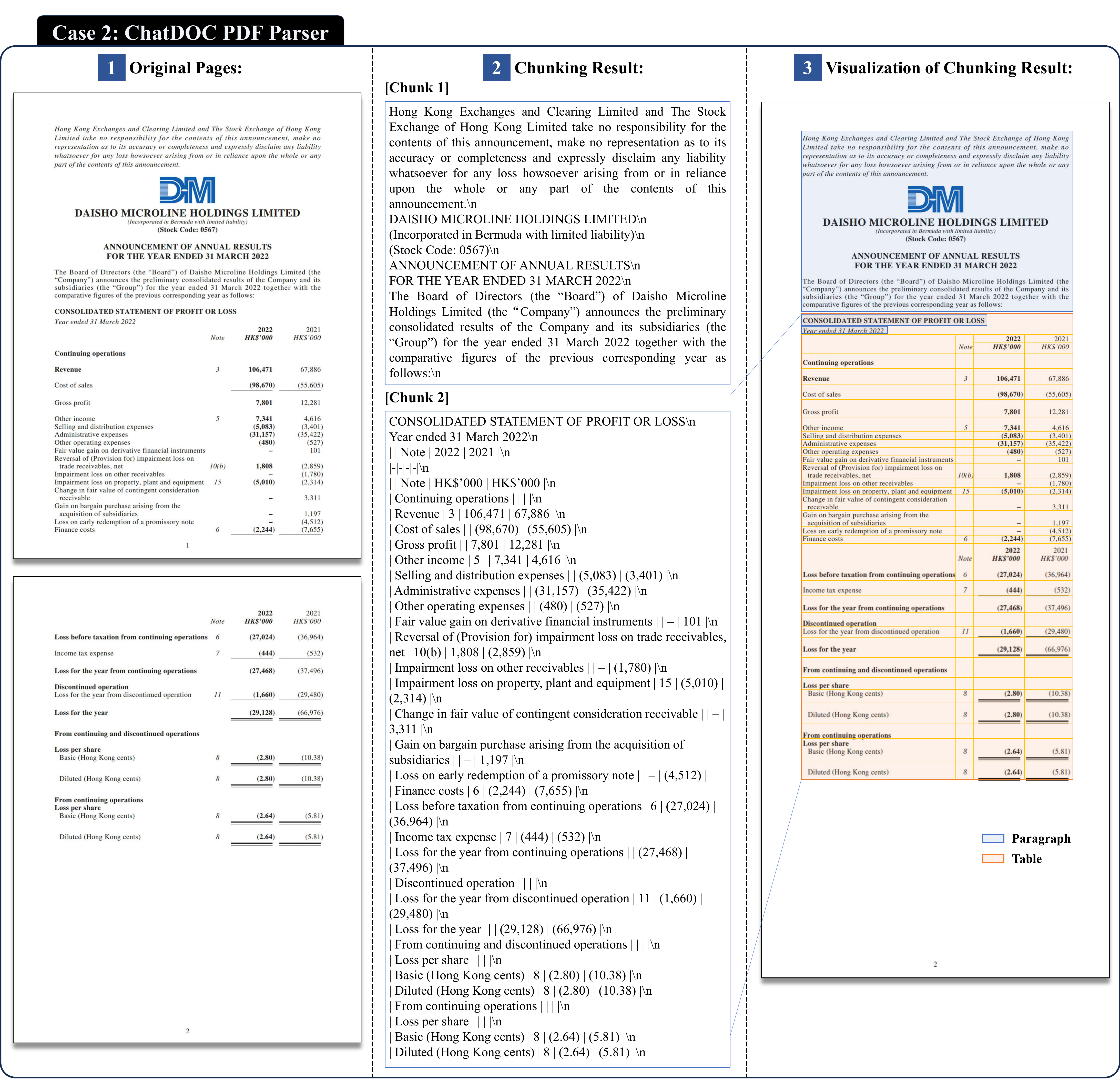}
\caption{Parsing and chunking result of ChatDOC PDF Parser on \nameref{par:case2} (original document: \cite{casedaisho}). Zoom in to see the details.}
\label{fig:case1chatdoc}
\end{figure}

For ChatDOC PDF Parser, shown in \ref{fig:case1chatdoc}, the parsing outcome is notably different. It not only preserves the document structure but also effectively segments the document in a way that maintains its inherent meaning. In this case, the table that spans two pages is set into one chunk, with its title at the beginning. So, the information in this chunk is self-contained. If this chunk is retrieved for RAG, the LLM can digest useful information within it.

\clearpage

\clearpage

\end{document}